\documentclass[10pt,twocolumn,letterpaper]{article}

\usepackage{cvpr}              

\usepackage{multirow}
\usepackage{subcaption}
\usepackage{amsmath}
\usepackage{booktabs}
\usepackage{amssymb}
\usepackage{algorithmic}
\usepackage{algorithm}
\usepackage{pifont}

\definecolor{cvprblue}{rgb}{0.21,0.49,0.74}
\usepackage[pagebackref,breaklinks,colorlinks,allcolors=cvprblue]{hyperref}
\usepackage{graphicx}
\usepackage{subcaption}
\usepackage{caption} 
\usepackage[accsupp]{axessibility} 

\def\paperID{10671} 
\def\confName{CVPR}
\def\confYear{2025}

\title{STAA-SNN: Spatial-Temporal Attention Aggregator for Spiking Neural Networks}

\author{
Tianqing Zhang\textsuperscript{\rm 1}\thanks{Equal contribution.} 
\quad
Kairong Yu\textsuperscript{\rm 1}\footnotemark[1] 
\quad 
Xian Zhong\textsuperscript{\rm 3}
\quad 
Hongwei Wang\textsuperscript{\rm 1}\thanks{Corresponding authors.}
\quad 
Qi Xu\textsuperscript{\rm 2}\footnotemark[2] 
\quad 
Qiang Zhang\textsuperscript{\rm 2}\\
\textsuperscript{\rm 1} Zhejiang University \hspace{8pt} 
\textsuperscript{\rm 2} Dalian University of Technology \hspace{8pt} 
\textsuperscript{\rm 3} Wuhan University of Technology\\
{\tt\small \textsuperscript{\rm 1}\{zhangtianqing,22271086\}@zju.edu.cn}
\hspace{5pt}
{\tt\small \textsuperscript{\rm 1}hongweiwang@intl.zju.edu.cn}\\
{\tt\small \textsuperscript{\rm 2}\{xuqi,zhangq\}@dlut.edu.cn}
\hspace{5pt}
{\tt\small \textsuperscript{\rm 3}zhongx@whut.edu.cn}
}

\begin{document}
\maketitle
\begin{abstract}
Spiking Neural Networks (SNNs) have gained significant attention due to their biological plausibility and energy efficiency, making them promising alternatives to Artificial Neural Networks (ANNs). However, the performance gap between SNNs and ANNs remains a substantial challenge hindering the widespread adoption of SNNs. 
In this paper, we propose a Spatial-Temporal Attention Aggregator SNN (STAA-SNN) framework, which dynamically focuses on and captures both spatial and temporal dependencies. 
First, we introduce a spike-driven self-attention mechanism specifically designed for SNNs. 
Additionally, we pioneeringly incorporate position encoding to integrate latent temporal relationships into the incoming features. 
For spatial-temporal information aggregation, we employ step attention to selectively amplify relevant features to variant steps. 
Finally, we implement a time-step random dropout strategy to avoid local optima. 
The framework demonstrates exceptional performance across diverse datasets and exhibits strong generalization capabilities. 
Notably, STAA-SNN achieves state-of-the-art results on neuromorphic datasets CIFAR10-DVS of 82.10\% and with performances of 97.14\%, 82.05\% and 70.40\% on the static datasets CIFAR-10, CIFAR-100 and ImageNet, respectively. 
Furthermore, this model exhibits improved performance ranging from 0.33\% to 2.80\% with fewer time steps. 
\end{abstract}    
\section{Introduction}
\label{sec:intro}
In recent years, Spiking Neural Networks (SNNs) have attracted significant attention due to their low energy consumption and biological interpretability\cite{roy2019towards,christensen20222022,wu2018spatio,wu2019direct,fang2021deep,wu2022brain}. 
Inspired by the behavior of biological neurons, SNNs represent information using discrete binary spikes over multiple timesteps, making them well-suited for implementation on low-power neuromorphic hardware, offering a distinct advantage over traditional Artificial Neural Networks (ANNs)\cite{akopyan2015truenorth,davies2018loihi,furber2014spinnaker,orchard2021efficient,yamazaki2022spiking}. 
Among various spiking neuron models, the Leaky Integrate-and-Fire (LIF) model\cite{abbott1999lapicque} is widely used in SNNs for its linear differentiable~\cite{lagani2023spiking,lee2020enabling}.
Although previous works\cite{hu2021advancing,zheng2021going,fang2021deep} have developed deep SNNs to accelerate model convergence, the multi-timestep computations inherent in SNNs significantly increase both training and inference times, leading to high latency. This limitation hinders the potential for efficient algorithm design.
Consequently, a performance gap still exists between directly trained large-scale SNNs and ANNs in pattern recognition tasks, underscoring the need for novel approaches that integrate insights from neuroscience with conventional deep learning to improve both the efficiency and accuracy of SNNs.
Inspired by the human visual and cognitive systems, attention mechanisms regulate neural activity and connectivity, enhancing responses to focal stimuli.
This enhancement is reflected in the increased firing rates and improved synchronization of neurons in relevant brain regions. Additionally, this mechanism influences the strength and efficiency of neural connections, thereby facilitating the flow of information within and between different brain areas.
In the field of deep learning, attention mechanisms enable neural networks to selectively focus on relatively important input information, leading to significant advancements in various domains such as natural language processing (NLP) and computer vision (CV). Compared to traditional architectures \cite{2021Is,2021Swin, 2021ViViT}, these mechanisms have consistently demonstrated superior performance.

In this paper, we explore the integration of attention mechanisms into deep SNNs. Although SNNs are biologically inspired models, their current implementations exhibit significant limitations due to inadequate biological fidelity. For instance, the heterogeneity among spiking neurons and the ability to independently transmit spatial and temporal information are overlooked, with membrane parameters being uniformly and permanently set across dimensions and layers in neural networks.
Neuroscience research \cite{turrigiano2004homeostatic, alcami2019beyond} has highlighted the critical role of synaptic homeostatic plasticity mechanisms in neuronal activity, as well as the regulatory function of glial cells in information transmission between neurons. Enhancing the alignment of SNN models with these biological processes has the potential to improve their overall performance. Therefore, we propose allowing different network layers to exhibit varying degrees of information transmission permeability.

We introduce a self-attention mechanism to capture spatial correlations and the significance of information, while integrating positional encoding to enhance spatial information and potentially learn temporal features. In traditional SNNs, input features are typically aggregated through simple addition, which can introduce significant noise into the synthesized features. To address this, we propose a Step Attention mechanism that aggregates and reconstructs features at each time step, thereby improving the quality of information propagation. Additionally, we introduce a novel Time-Step Random Dropout (TSRD) strategy to prevent the network from falling into local optima, thereby enhancing its robustness and accelerating training.
Our main contributions are summarized as follows:
\begin{itemize}
\item Inspired by neuroscience and attention mechanisms, we propose a novel Spatio-Temporal Attention Aggregator for Spiking Neural Networks (STAA-SNN) that integrates self-attention, positional encoding, and step attention into a unified framework, enabling SNNs to efficiently extract features across temporal and spatial domains while maintaining stable training.
\item We introduce a Time-Step Random Dropout (TSRD) strategy to accelerate training and enhance model generalization by preventing premature convergence and enabling optimal performance.
\item We evaluate the performance, convergence, and sparse spiking activity of STAA-SNN across multiple benchmark datasets, demonstrating that our method achieves state-of-the-art or superior results with fewer timesteps, highlighting its efficiency and effectiveness.
\end{itemize}



\begin{figure*}[!h]
    \centering
    \includegraphics[width=\linewidth]{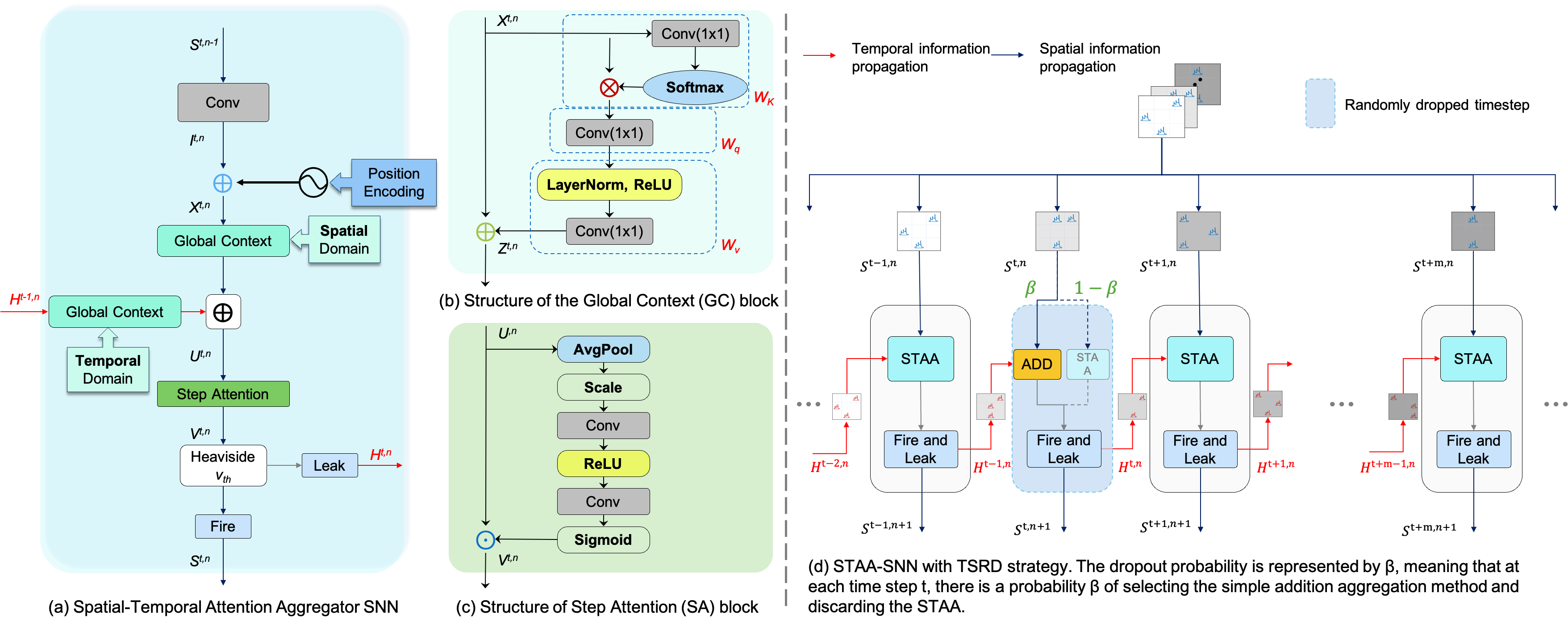}
    \caption{Overview of the STAA-SNN Architecture and the TSRD strategy.}
    \label{fig:overview}
\end{figure*}

\section{Related Works}

\subsection{Training of Deep SNNs}

Training deep SNNs generally follows two primary approaches: ANN-to-SNN conversion and direct SNN training.
The ANN-to-SNN conversion method aims to transfer knowledge from traditional ANNs to SNNs, but it often encounters challenges such as minimizing accuracy degradation while optimizing inference latency and energy efficiency\cite{han2020deep,wang2023new,han2020rmp}. 
Moreover, this approach struggles to fully leverage the spatiotemporal dynamics inherent in SNNs due to the lack of temporal encoding in ANNs.
Direct SNN training is categorized into unsupervised and supervised learning. Unsupervised methods rely on biologically plausible learning rules such as Hebbian learning\cite{hebb2005organization} and Spike-Timing-Dependent Plasticity (STDP)\cite{bi1998synaptic}. However, these approaches are generally not suitable for deep SNNs or large-scale datasets. On the other hand, supervised learning integrates backpropagation with surrogate gradients\cite{neftci2019surrogate,lee2020enabling,wu2019direct,fang2021deep}, making applicability increasingly viable\cite{xu_biologically_2023,xu_hybrid_2024}.
Several methods have been proposed to enhance direct SNN training. 
For instance, MPBN\cite{guo2023membrane} incorporates a batch normalization layer after membrane potential updates to stabilize training, while MPD-AL\cite{zhang2019mpd} introduces an alternative learning paradigm that adapts synaptic modifications based on membrane potential traces. MS-ResNet\cite{hu2021advancing} reorganizes Vanilla ResNet layers to better align with SNNs, improving feature extraction and gradient propagation. Furthermore, IM-Loss\cite{guo2022loss} optimizes information maximization loss to increase spike information entropy. Recently, RSNN\cite{xu2024rsnn} introduced a Convolutional LSTM-based approach that preserves temporal correlations, enhancing spike-based data processing in SNNs.

\subsection{Attention Mechanism in SNNs}
Attention mechanisms dynamically assign varying levels of importance to different input components, thereby facilitating efficient processing and feature extraction. 
Since their introduction in 2014, attention mechanisms have become integral to enhancing deep learning models and have been widely applied to traditional ANNs\cite{Hu_2018_CVPR}.
In deep learning, attention mechanisms are primarily used in two ways: as a basic paradigm for conducting meta-operator such as self-attention~\cite{2017Attention} or as an auxiliary enhancement modules~\cite{2020Squeeze}. Also these two approaches can be combined.
Recently, integrating attention mechanisms into SNNs has gained traction as a means to enhance their performance. 
Unlike traditional ANNs, where attention typically refines spatial feature representations, SNN-based attention mechanisms must consider both spatial and temporal information. Early work by \cite{neokleous2011selective} introduced a computational model that bridges human selective attention and conscious awareness.
Kundu$\eta$~\cite{kundu2021spike} utilize attention mechanisms to compress SNN models.
Temporal-wise attention was first introduced in TA-SNN\cite{yao2021temporal}, laying the foundation for subsequent developments in SNN-specific attention mechanisms. 
MA-SNN \cite{yao2022attention} proposed temporal-channel-spatial attention for SNNs, surpassing models focused on single-dimensional information. 
TCJA-SNN\cite{zhu2024tcja} introduced a temporal-channel joint attention mechanism, while SCTFA \cite{cai_spatialchanneltemporal-fused_2023} applied temporal-channel-spatial attention to SNNs with promising results. 
Moreover, STSC-SNN \cite{yu_stsc-snn_2022} incorporated temporal convolution and attention mechanisms to implement synaptic filtering and gating functions, demonstrating the versatility of attention mechanisms in enhancing SNN capabilities. 
IM-LIF \cite{lian_im-lif_2024} adopted temporal-wise attention to improve the updating of GRUs and LSTMs within SNNs.
More recently, researchers have explored incorporating SNNs into Transformer architectures, achieving promising results on multiple benchmark datasets\cite{she2022spikeformer,li2022spikeformer}.
These researches reveal that attention mechanisms can significantly improve the performance of SNN. 
However, there remains a need for more specialized and high-performance attention modules tailored explicitly for SNN architectures. Addressing this gap can further enhance the efficiency and robustness of SNNs in real-world applications.



\section{Methodology}

In this section, we will first provide a brief overview of the Leaky Integrate-and-Fire (LIF) model and the training process.
Following that, we introduce the proposed Spatial-Temporal Attention Aggregator.
Lastly, we outline the timestep random dropout strategy for SNN.

\subsection{Leaky Integrate-and-Fire Model}
The spiking neuron is a fundamental component of SNNs. 
In this work, we adopt the Leaky Integrate-and-Fire (LIF) neuron due to its efficiency and simplicity. 
The discrete-time, iterative update rule of the LIF model is given by:
\begin{align}
    & V^{t,n} = H^{t-1,n} + \frac{1}{\tau} [I^{t-1,n} - (H^{t-1,n} - V_\text{reset})],
    \label{eq:bad_lif} \\
    & S^{t,n} = \Theta(V^{t,n} - v_\text{th}), \\
    & H^{t,n} = V_\text{reset} \cdot S^{t,n} + V^{t,n} \odot (1 - S^{t,n}).
\end{align}
Here, \(H^{t-1,n}\) denotes the membrane potential after a spike trigger at the previous timestep, while \(I^{t,n}\) and \(V^{t,n}\) represent the input and membrane potential of the \(n\)-th layer at timestep \(t\), respectively.
The threshold \(v_{th}\) determines whether a spike is emitted, producing the binary spike sensor
\(S^{t,n}\).
The Heaviside step function \(\Theta\) outputs 1 if $x \geq 0$ and 0 otherwise.

Given the similarity between iterative LIF dynamics and Recurrent Neural Networks (RNNs), Spatio-Temporal Backpropagation (STBP)~\cite{wu2018spatio,wu2019direct} enables direct SNN training via Backpropagation Through Time (BPTT):

\begin{equation}
    \resizebox{0.95\linewidth}{!}{$
            \displaystyle
            \frac{\partial L}{\partial W^n} = 
            \sum \limits_t \left(
            \frac{\partial L}{\partial S^{t,n}} \cdot \frac{\partial S^{t,n}}{\partial V^{t,n}} + \frac{\partial L}{\partial V^{t+1,n}} \cdot \frac{\partial V^{t+1,n}}{\partial V^{t,n}}
             \right)
            \frac{\partial V^{t,n}}{\partial W^n}
        $}.
\end{equation}
However, the term $\frac{\partial S^{t,n}}{\partial V^{t,n}}$, which represents the derivative of the spike function, is non-differentiable. 
To address this, prior works introduce Surrogate Gradients(SG), commonly approximated using a rectangular function:

\begin{equation}
    \frac{\partial S^t}{\partial V^t} = 
    \frac{1}{a} \cdot \text{sign}(|V^{t}-V_\text{th}|< \frac{a}{2}),
\end{equation}
where \(a\) is a hypeparameter and set to 1.
This approximation provides a gradient of 1 within the range \(V_\text{th}-0.5 \leq V^{t} \leq V_\text{th}+0.5\), and otherwise 0.

\paragraph{\textbf{Neuromorphic-Inspired Adaptive LIF Design.}}
Examining~\cref{eq:bad_lif}, it is evident that keeping $V_\text{reset}$ and $\tau$ constant results in fixed weight coefficients for \(H\) and \(I\), which remain unchanged across timestep  \(t\) and layer \(n\). 
This simplification deviates from biological neural processing, where neuronal excitability dynamically adapts to stimuli.
To bridge this gap, we propose an adaptive modification to the membrane potential update rule:

\begin{equation}
    \label{eq:now_lif}
    V^{t,n} = M \odot H^{t-1,n} + N \odot I^{t-1,n}.
\end{equation}
Here, \(M\) and \(N\) are coefficient matrices that dynamically adjust through iterative learning, regulating activation and inhibition probabilities across layers. This biologically inspired adaptation enhances information processing efficiency and robustness in SNNs.

\paragraph{\textbf{Global Context (GC) Block.}}
The Global Context (GC) Block serves as a lightweight alternative to the self-attention mechanism used in Transformers, enabling the integration of self-attention into conventional convolution-based deep learning architectures without the computational overhead of full Transformer models. This makes it possible to enhance feature learning in non-Transformer structures while maintaining efficiency and precision.
In this paper, we simulate the self-attention operation using a GC block \cite{cao2019gcnet}, which consists of three $1 \times 1$ convolutional kernels corresponding to key, query, and value. 
As depicted in~\cref{fig:overview}(b), 
the spatial spiking input tensor of the \(n\)-th layer at the \(t\)-th timestep is defined as $X^{t,n} \in \mathbb{R}^{C^n \times H^n \times W^n}$, where \(W^n\) and \(H^n\) denote the width and height of the feature map, and \(C^n\) represents the number of channels. 
Passing \(X^{t,n}\) through the GC block yields three weight matrices for self-attention, namely \(W_k\), \(W_q\) and \(W_v\), defined as:
\begin{align}
    W_k & = X^{t,n} \cdot \text{Sigmoid}(\text{Conv}_k(X^{t,n})), \\
    W_q & = \text{Conv}_q(W_k), \\
    W_v & = \text{Conv}_v(\text{ReLU}(\text{LN}(W_q))).
\end{align}
Before computing $W_k$, $X^{t,n}$ is reshaped from $C \times H \times W$ to $C \times HW$. The resulting matrices $W_k$, $W_q$, $W_v \in \mathbb{R}^{C \times 1 \times 1}$ encode self-attention weights, which are then used to refine the input feature map. 
Subsequently, the enhanced output is computed as:
\begin{equation}
    Z^{t,n} = X^{t,n} \oplus W_v,
\end{equation}
where \( Z^{t,n} \in \mathbb{R}^{C, H, W} \) maintains the same dimensions as \(X^{t,n}\).
The refined feature map \(Z^{t,n}\), in comparison to the \(X^{t,n}\), improves the network’s ability to focus on self-relevant and task-critical information while suppressing irrelevant signals. 
This approach facilitates the seamless transmission of enriched data flow, enabling the retention of valuable features for subsequent stages.

\begin{table*}[]
\small
\centering
\begin{tabular}{ccccll}
\toprule
 & \multirow{2}{*}{\textbf{Method}}           & \multirow{2}{*}{\textbf{Architecture}} & \multirow{2}{*}{\textbf{Timestep}} & \multicolumn{2}{c}{\textbf{Accuracy (\%)}}                                 \\
\cmidrule(lr){5-6} 
 &   &  &  & \multicolumn{1}{c}{\textbf{CIFAR-10}} & \multicolumn{1}{c}{\textbf{CIFAR-100}} \\
\midrule
\multirow{20}{*}{\textbf{CNN-based}} &\multirow{2}{*}{GLIF~\cite{yao2022glif}}             & ResNet-19                                         & 2                                             & 94.44\%                        & 75.48\%                         \\&
                                  & ResNet-19                                         & 4                                             & 94.85\%                        & 77.05\%                         \\
\cmidrule(lr){2-6} 
&  \multirow{2}{*}{TET~\cite{deng2022temporal}}              & ResNet-19                                         & 2                                             & 94.16\%                        & 72.87\%                         \\&
                                  & ResNet-19                                         & 4                                             & 94.44\%                        & 74.47\%                        \\
                                  \cmidrule(lr){2-6} &
\multirow{2}{*}{LSG~\cite{lian2023learnable}}              & ResNet-19                                         & 2                                             & 94.41\%                        & 76.32\%                         \\&
                                  & ResNet-19                                         & 4                                             & 95.17\%                        & 76.85\%                         \\
                                  \cmidrule(lr){2-6} &
\multirow{2}{*}{PFA~\cite{deng2024tensor}}              & ResNet-19                                         & 2                                             & 95.6\%                         & 76.7\%                         \\&
                                  & ResNet-19                                         & 4                                             & 95.71\%                        & 78.1\%                         \\
                                  \cmidrule(lr){2-6} &
\multirow{2}{*}{Diet-SNN~\cite{rathi2021diet}}         & ResNet-20                                         & 5                                             & 91.78\%                        & 64.07\%                         \\&
                                  & VGG-16                                            & 5                                             & 93.85\%                        & 69.67\%                         \\
                                  \cmidrule(lr){2-6} &
IM-loss~\cite{guo2022loss}       & VGG-16                                            & 5                                             & 93.85\%                        & 70.18\%                        \\
\cmidrule(lr){2-6} &
IM-LIF~\cite{lian_im-lif_2024}        & ResNet-19                                         & 3                                             & 95.29\%                        & 77.21\%                         \\
\cmidrule(lr){2-6} & \multirow{6}{*}{MPBN~\cite{guo2023membrane}}             & ResNet-19      & 1                                             & 96.06\%                        & 78.71\%                         \\ &
                                  & ResNet-19                                         & 2                                             & 96.47\%                        & 79.51\%                         \\&
                                  & ResNet-19                                         & 4                                             & 96.52\%                        & 80.1\%                          \\&
                                  & ResNet-20                                         & 1                                             & 92.22\%                        & 68.41\%                         \\&
                                  & ResNet-20                                         & 2                                             & 93.54\%                        & 70.79\%                         \\&
                                  & ResNet-20                                         & 4                                             & 94.28\%                        & 72.3\%                          \\
\midrule
\multirow{4}{*}{\textbf{Transformer-based}} &
Spikformer\cite{zhou2022spikformer}    & Spikformer-4-25                                   & 4                                             & 93.94\%                        & 75.96\%                         \\
\cmidrule(lr){2-6} 
& Spikingformer\cite{zhou2023spikingformer} & Spikingformer-4-25                                & 4                                             & 94.77\%                        & 77.43\%                         \\
\cmidrule(lr){2-6}
& Spike-driven Transformer~\cite{yao2024spike} & Transformer-2-512 & 4 & 95.60\% & 78.40\%\\ 
\midrule
\multirow{8}{*}{\textbf{CNN-based}}&  \multirow{8}{*}{\textbf{Ours}}             & VGG-13                                            & 2                                             & 94.70\%{\tiny$\pm$ 0.12\%}  & 75.16\%{\tiny$\pm$ 0.08\%}  \\
&                                  & VGG-13                                            & 4                                             & 95.26\%{\tiny$\pm$ 0.09\%}  & 76.35\%{\tiny$\pm$ 0.11\%}  \\
&                                  & ResNet-19                                         & 1                                             & 96.75\%{\tiny$\pm$ 0.09\%} & 79.37\%{\tiny$\pm$ 0.10\%} \\
&                                  & ResNet-19                                         & 2                                             & 96.85\%{\tiny$\pm$ 0.11\%}  & 80.57\%{\tiny$\pm$ 0.11\%}  \\
&                                  & ResNet-19                                         & 4                                             & \textbf{97.14\%}{\tiny$\pm$ 0.10\%} & \textbf{82.05\%}{\tiny$\pm$ 0.10\%}  \\
&                                  & ResNet-20                                         & 1                                             & 93.08\%{\tiny$\pm$ 0.09\%}  & 70.14\%{\tiny$\pm$ 0.07\%}  \\
&                                  & ResNet-20                                         & 2                                             & 94.35\%{\tiny$\pm$ 0.08\%}  & 73.20\%{\tiny$\pm$ 0.12\%}  \\
&                                  & ResNet-20                                         & 4                                             & 95.03\%{\tiny$\pm$ 0.11\%}  & 75.10\%{\tiny$\pm$ 0.12\%}  \\
\bottomrule
\end{tabular}
\caption{Comparison results with SOTA methods on CIFAR-10/100.}
\end{table*}

\paragraph{\textbf{Position Encoding (PE) Block.}}

Position encoding is widely used in NLP and sequence modeling to incorporate positional information into input data. While SNNs share structural similarities with RNNs and inherently process temporal information, their reliance on sparse binary spikes for information storage can lead to significant feature loss. This study introduces position encoding to enhance spatial features while enriching temporal relationships, thereby improving feature retention and strengthening temporal representations.
The positional encoding is computed as follows:
\begin{equation}
    X^{t,n} = I^{t,n} \oplus \text{Pos}_{t,n},
\end{equation}
where \(\text{Pos}_{t,n}\) represents the position encoding at the \(t\)-th time step in the \(n\)-th layer, with \(\text{Pos}_{t,n} \in \mathbb{R}^{T, C^n}\).

\paragraph{\textbf{Step Attention (SA) Block.}}
While previous modules primarily enhance feature processing from a spatial perspective, SNNs uniquely process information across both spatial and temporal. 
Unlike other types of network, SNNs offer an additional understanding of information sensitivity across different timesteps. Therefore, in this paper, we introduce Step Attention to tackle the temporal challenges inherent in SNNs.
As illustrated in ~\cref{fig:overview}(c), let \(U^{t,n} \in \mathbb{R}^{C^n \times H^n \times W^n}\) denote the spatially aggregated feature input at the \(n\)-th layer and the \(t\)-th timestep.
The specific Step Attention mechanism is defined as:
\begin{equation}
    \begin{split}
        V^{t,n} &= U^{t,n} \odot \text{Sigmoid}\bigg( \text{Conv}_2 \bigg( \text{ReLU} \bigg( \text{Conv}_1 \\
        & \quad \big( \alpha \cdot \text{AvgPool}(U^{t,n}) \big) \bigg) \bigg) \bigg),
    \end{split}
\end{equation}
where \(\alpha\) is a scaling factor that regulates the initial training speed for smoother convergence. In practice, \(\alpha\) is typically set to a default value of 2.

\paragraph{\textbf{LIF with Spatio-Temporal Attention Aggregator (STAA).}}
The STAA seamlessly integrates with the LIF model, offering compatibility across various network structures without constraints. It adeptly harnesses the potential of spiking neurons, effectively optimizing the storage of discrete spike information while minimizing the loss of crucial data during transmission.
As shown in~\cref{fig:overview}(a), the neuronal update equation for the LIF model with STAA is formulated as:
\begin{align}
    U^{t,n} & = GC_1(X^{t,n}) \oplus GC_2(H^{t-1,n}), \\
    V^{t,n} & = SA(U^{t,n}).
\end{align}
Where $GC_1$ and $GC_2$ represent Global Context Blocks for the input $X^{t,n}$ and the previous membrane state $H^{t-1,n}$, respectively. 
The remaining update equations follow the original LIF model.

\subsection{Time Step Random Dropout (TSRD) Strategy}\label{sec:tsrd}

To further enhance training efficiency, we propose TSRD, an adaptive training strategy designed to prevent premature feature stabilization in deep SNNs. 
In deeper timesteps, spiking features may solidify too early during iterations, hindering the network from achieving optimal convergence.
As illustrated in~\cref{fig:overview}(d),
TSRD randomly dropout the augmentation module at timestep \(t \in (0, T)\) with a dropout probability \(\beta\), and instead choose the most basic element-wise addition aggregation method.
The TSRD technique accelerates training and helps the network escape local optima, thereby improving generalization performance.
\begin{table}[h]
\centering
\small
\setlength{\tabcolsep}{4pt}
\begin{tabular}{cccc}
\toprule 
\textbf{Method}                         & \textbf{Architecture}       & \textbf{Timestep}   & \textbf{Accuracy}         \\ 
\midrule
STBP-tdBN ~\cite{zheng2021going}                       & ResNet34          & 6          & 63.72\%          \\
\midrule
TET~\cite{deng2022temporal}                           & ResNet34          & 6          & 64.79\%          \\
\midrule
RecDis-SNN~\cite{guo_recdis-snn_2022}                    & ResNet34          & 6          & 67.33\%          \\
\midrule
GLIF~\cite{yao2022glif}                           & ResNet34          & 4          & 67.52\%          \\
\midrule
IM-Loss~\cite{guo2022loss}                      & ResNet18          & 6          & 67.43\%          \\ 
\midrule
\multirow{2}{*}{Real Spike~\cite{guo2022real}}    & ResNet18          & 4          & 63.68\%          \\
                               & ResNet34          & 4          & 67.69\%          \\ 
\midrule
\multirow{2}{*}{RMP-Loss~\cite{guo_rmp-loss_2023}}      & ResNet18          & 4          & 63.03\%          \\
                               & ResNet34          & 4          & 65.17\%          \\ \midrule
\multirow{2}{*}{MPBN~\cite{guo2023membrane}}          & ResNet18          & 4          & 63.14\%          \\
                               & ResNet34          & 4          & 64.71\%          \\ \midrule
\multirow{2}{*}{SEW ResNet~\cite{fang2021deep}}    & ResNet18          & 4          & 63.18\%          \\
                               & ResNet34          & 4          & 67.04\%          \\ \midrule
\multirow{2}{*}{\textbf{Ours}} & ResNet18 & 4 &\textbf{68.27\%}{\tiny$\pm$ 0.19\%} \\
                               & ResNet34 & 4 &\textbf{70.40\%}{\tiny$\pm$ 0.15\%} \\ \bottomrule
\end{tabular}
\caption{Comparison of training based SNN SOTA on ImageNet.}
\label{tab:imagenet}
\end{table}

\begin{table}
    \centering
    \small
    \setlength{\tabcolsep}{2.5pt}
    \begin{tabular}{cccc}
        \toprule
         \textbf{Methods} & \textbf{Architecture} & \textbf{Timestep} & \textbf{Accuracy} \\
         \midrule
         \multicolumn{4}{c}{\textbf{CIFAR10-DVS}} \\
         \midrule
         IM-loss~\cite{guo2022loss} & ResNet-19 & 10 & 72.60\% \\
         LSG~\cite{lian2023learnable} & ResNet-19 & 10 & 77.90\% \\
         MPBN~\cite{guo2023membrane} & ResNet-19 & 10 & 74.40\% \\
         MPBN~\cite{guo2023membrane} & ResNet-20 & 10 & 78.70\% \\
         TET~\cite{deng2022temporal} & VGGSNN & 10 & 77.30\% \\
         IM-LIF~\cite{lian_im-lif_2024} & VGG-13 & 10 & 80.50\% \\
         GLIF~\cite{yao2022glif} & 7B-wideNet & 16 & 78.10\% \\
         STSA~\cite{wang_spatial-temporal_2023} & STS-Transformer & 16 & 79.93\% \\
         SEW~\cite{fang2021deep} & SEW-ResNet & 16 & 74.4\% \\
         Spikeformer~\cite{li2022spikeformer} & Spikeformer & 16 & 80.9\% \\
         \midrule
         \multirow{2}{*}{\textbf{Ours}} & ResNet-20 & 16 & 81.90\%\tiny{$\pm$ 0.20\%}   \\
                                        & VGG-13    & 16 & \textbf{82.10\%}\tiny{$\pm$ 0.20\%}  \\
         \midrule
         \multicolumn{4}{c}{\textbf{DVS128 Gesture}} \\
         \midrule
         STBP-tdBN~\cite{zheng2021going} & ResNet-17 & 40 & 96.87\% \\
         SEW~\cite{fang2021deep} & 7B-Net & 16 & 97.52\% \\
         PLIF~\cite{fang2021incorporating} & PLIFNet & 20 & 97.57\% \\
         MA-SNN~\cite{yao2022attention} & 5 layers SCNN & 20 & 98.23\% \\
         ASA-SNN~\cite{yao_inherent_2023} & 5 layers SCNN & 20 & 97.70\% \\
         IM-LIF~\cite{lian_im-lif_2024} & ResNet-19 & 40 & 97.33\% \\
         LIAF+TA~\cite{yao_temporal-wise_2021} & TA-SNN-Net & 60 & 98.61\% \\
         Spike-driven- & \multirow{2}{*}{Transformer-2-512}    & \multirow{2}{*}{16} & \multirow{2}{*}{\textbf{99.30\%}} \\ 
          Transformer~\cite{yao2024spike} & &  \\
         \midrule
         \textbf{Ours} & VGG-13 & 16 & 98.61\%\tiny{$\pm$ 0.20\%}  \\
         \bottomrule
    \end{tabular}
    \caption{Comparison results with SOTA methods on CIFAR10-DVS and DVS128 Gesture.}
    \label{tab:exp_dvs}
\end{table}

\section{Experiments}
We first outlined the experimental setup, including dataset details and implementation specifics. We then compare our results with previous SOTA methods across multiple benchmarks. Next, we perform ablation studies to assess the effectiveness of STAA. 
Finally, we visualize the model’s performance.
Further details on datasets, hyperparameters, additional experiments, and computational efficiency analysis are provided in the \textbf{Supplementary Material}.

\begin{figure*}[t]
    \centering
    \includegraphics[width=\textwidth]{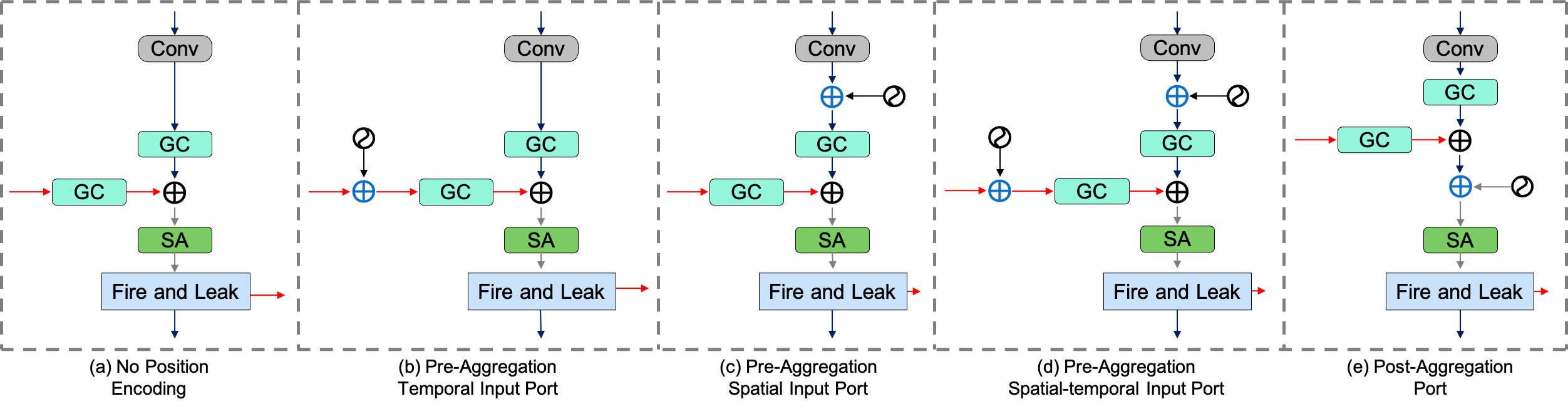}
    \caption{Position encoding locations in SNNs.
    }
    \label{fig:abla_port}
\end{figure*}

\subsection{Experimental Setup}
\label{sec:exp_setup}
\paragraph{\textbf{Datasets.}}

We evaluate our method on five benchmark datasets. 
CIFAR-10/100\cite{krizhevsky2010cifar} are standard image classification datasets with 50,000 training and 10,000 test images across 10 and 100 classes, respectively. ImageNet\cite{deng_imagenet_2009} is a large-scale dataset containing 1.2 million training, 50,000 validation, and 100,000 test images spanning 1,000 categories.
For neuromorphic vision tasks, CIFAR10-DVS\cite{li2017cifar10,lei2024motion,liu2024line} consists of 10,000 event streams across 10 classes, with a 9:1 training-to-testing split. DVS128 Gesture\cite{amir2017low} is an event-based gesture recognition dataset with 1,176 training and 288 test samples, featuring 11 gesture classes recorded from 29 subjects under three lighting conditions.

\paragraph{\textbf{Implementation Details.}}
In our experiments, we set the firing threshold \(v_\text{th}\) to 1, and initialize $\alpha$ and $\beta$ to 2 and 0.1, respectively. 
In the PE block, we adopt a learnable position encoding method, initializing all values to zero.
All experiments are implemented in PyTorch and conducted on an NVIDIA RTX 3090 GPU, except for ImageNet, which is trained on 8 NVIDIA RTX 4090 GPUs. 
The total training epochs are 500 for CIFAR-10, CIFAR-100, and CIFAR10-DVS, while ImageNet runs for 300 epochs.


\subsection{Performance Comparison}

\paragraph{\textbf{Static Image Classification.}}
We evaluate our model on three static datasets: CIFAR-10, CIFAR-100, and ImageNet. 
For CIFAR-10/100, STAA-SNNs are applied to both ResNet-like and VGG-like networks and tested at 1, 2, 3, and 4 timesteps. 
On ImageNet, the model is evaluated using ResNet-18 and ResNet-34 with 4 time steps.
Our method achieves top accuracies of 97.14\% on CIFAR-10, 82.05\% on CIFAR-100, and 70.40\% on ImageNet, surpassing previous state-of-the-art methods across various architectures. 
These improvements are attributed to the strong spatial-temporal attention mechanism and feature aggregation capabilities of STAA-SNNs, which fully leverage the potential performance of SNNs. Furthermore, our model achieves equivalent accuracy while requiring at least one fewer timestep compared to prior methods, improving efficiency without sacrificing performance.

\paragraph{\textbf{Event-based Action Recognition.}}
To further assess the spatial-temporal processing capabilities of STAA-SNNs, we conduct experiments on the DVS128 Gesture and CIFAR10-DVS datasets. Unlike static datasets, these neuromorphic datasets contain temporal event streams, requiring models to efficiently process spiking sequences.
As shown in \cref{tab:exp_dvs}, our model achieves state-of-the-art or superior performance with a smaller network and fewer timesteps. Specifically, STAA-SNNs achieve 82.10\% accuracy on CIFAR10-DVS and 98.61\% on the DVS128 Gesture dataset, demonstrating their effectiveness in event-based action recognition tasks.

\subsection{Ablation Study}

All ablation studies with T=4 using ResNet-20 as baseline.

\label{sec:exp_ablation}

\begin{figure}[]
    \centering
    \includegraphics[width=0.9\columnwidth]{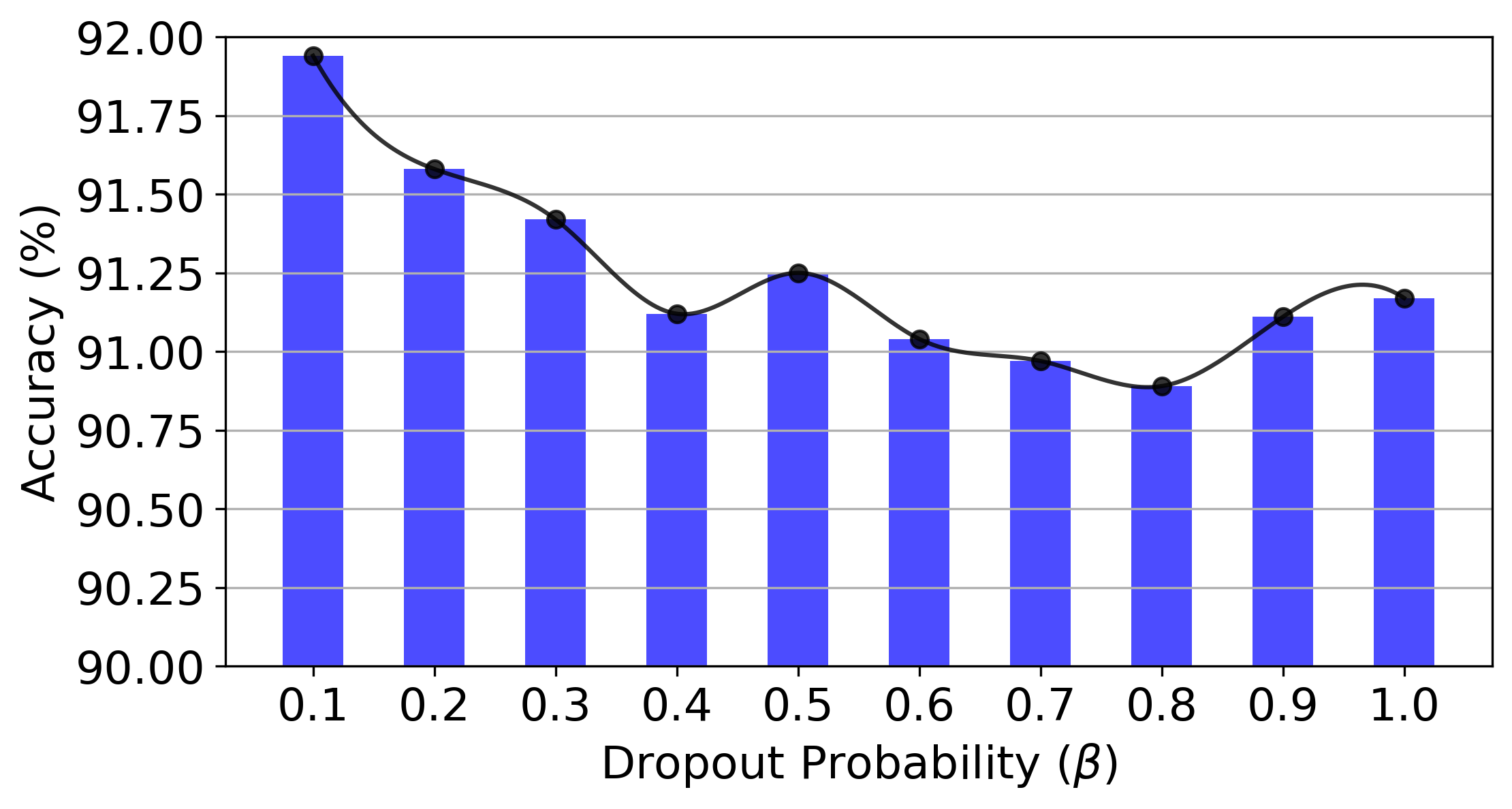}
    \caption{Distribution of accuracy with different dropout probability $\beta$ in TSRD on CIFAR-10.}
    \label{fig:abl_beta}
\end{figure}

\begin{table}[t]
\small
\centering
\begin{tabular}{cccccc}
\toprule
\multicolumn{3}{c}{STAA Components} & \multirow{2}{*}{TSRD} & \multirow{2}{*}{Accuracy} & \multirow{2}{*}{Accuracy Gain}  \\ 
\cmidrule{1-3}
GC          & PE           & SA   &      \\ \midrule
\ding{55}  &\ding{55}   & \ding{55}   & \ding{55}    & 72.30\%    & - \\
\checkmark &\ding{55}   & \ding{55}   & \ding{55}    & 73.22\%    & +0.92\% \\
\checkmark & \checkmark & \ding{55}   & \ding{55}    & 73.79\%    & +1.49\%  \\   
\checkmark & \checkmark & \ding{55}   & \checkmark    & 73.96\%    & +1.66\%  \\  
\checkmark & \checkmark & \checkmark & \ding{55}    & 74.78\%    & +2.48\%  \\
\checkmark & \checkmark    & \checkmark  & \checkmark   & 75.10\%    & +2.80\%  \\
\bottomrule
\end{tabular}
\caption{Combinations of sub-modules in STAA with TSRD.}
\label{tab:comb_staa}
\end{table}

\begin{figure*}[t]
    \centering
    \includegraphics[width=0.85\linewidth]{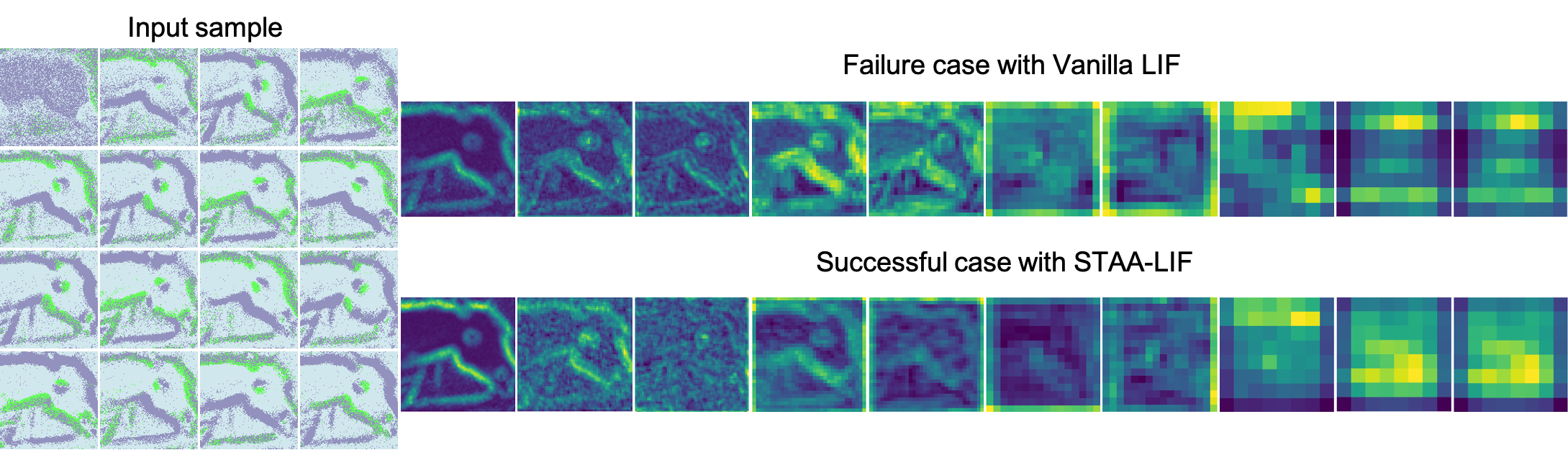}
    \caption{Visualization on CIFAR10-DVS. 
    Ten layers from VGG-13 in a shallow to deep manner.
    }
    \label{fig:heatmap}
\end{figure*}

\begin{figure}
    \centering
    \includegraphics[width=0.85\linewidth]{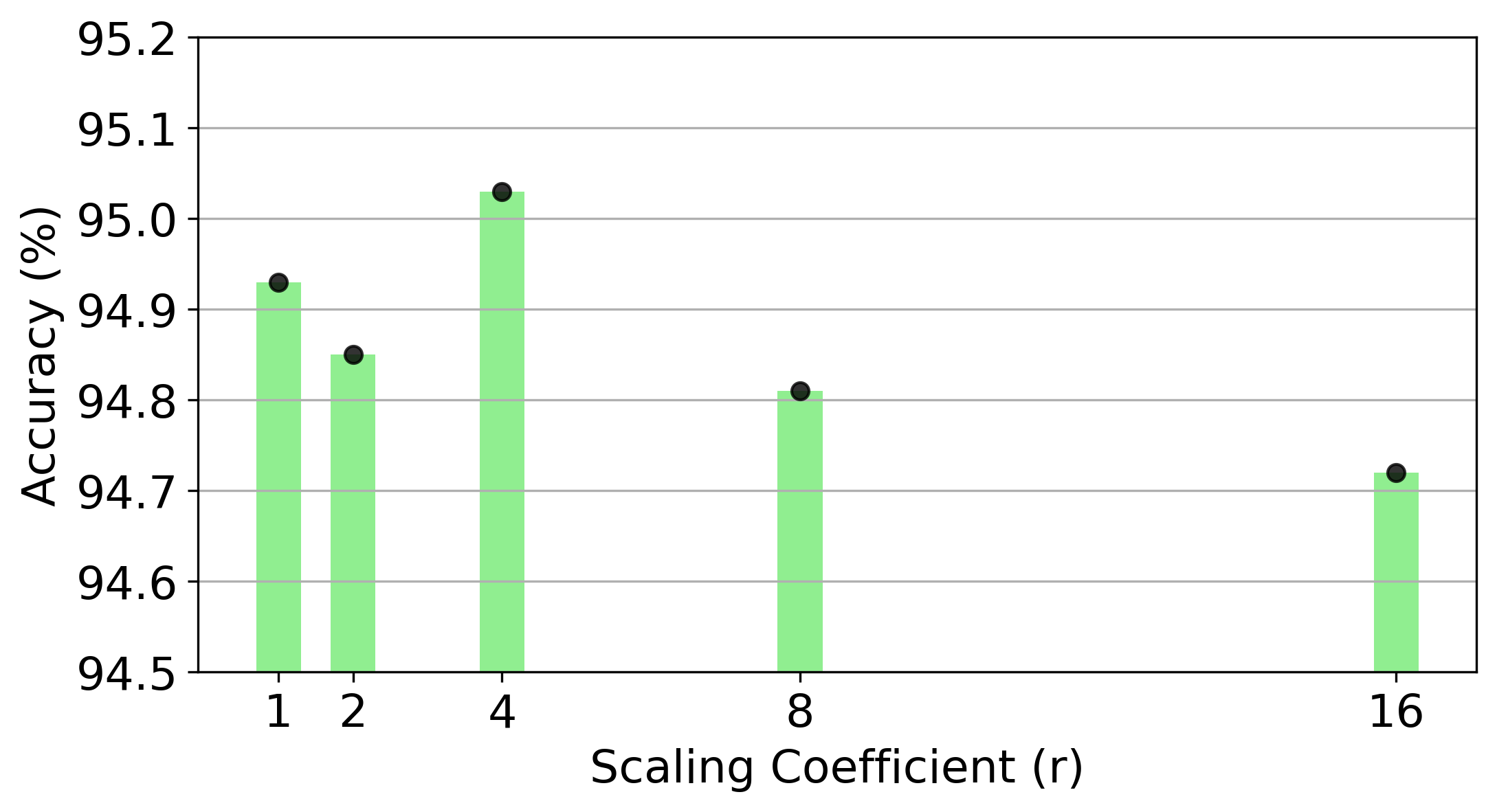}
    \caption{Impact of different scaling coefficients \(r\) for intermediate feature dimensions in the GC module on the CIFAR-10 dataset.}
    \label{fig:scale}
\end{figure}

\paragraph{Position encoding placement.}
To optimize PE in SNNs, we evaluate its best placement within the network. Given the structural complexity of SNNs compared to CNNs, identifying the most effective integration point is crucial.
As shown in \cref{fig:abla_port}, we evaluate four placement strategies (b)-(e) alongside a baseline SNN without PE (a), ensuring all other conditions remain constant. 
The CIFAR-10 accuracies for (b) to (e) are 94.08\%, 94.29\%, 94.27\%, and 93.96\%, respectively. The highest performance occurs when PE is placed at the pre-aggregation spatial input port (c). Although (b) and (d) yield similar results, (d) introduces additional parameters without significant gains. Thus, placing PE at the pre-aggregation spatial input port achieves the best balance between accuracy and efficiency.

\paragraph{Evaluation of position encoding methods.}
We next evaluate different encoding methods to determine the most effective approach. 
The results on CIFAR-10 indicate that learnable PE achieves an accuracy of 94.99\%, outperforming fixed PE, which attains 94.42\%. This 0.57\% improvement highlights the advantage of adaptive position encoding, as it enables the network to dynamically adjust spatial representations, leading to better feature alignment and overall performance.

\paragraph{Impact of dropout probability \(\beta\) in TSRD.}
To accelerate training and enhance generalization in TSRD, it is essential to determine an optimal dropout probability \(\beta\). Excessive dropout may lead to network collapse by discarding too much information, while insufficient dropout may limit its regularization effect.
As shown in \cref{fig:abl_beta}, by incrementally increasing \(\beta\) from 0.1 to 1.0, the model achieves relatively high performance when \( \beta \leq 0.3\), and achieving optimal at 0.1. These results confirm that excessive dropout removes essential information, degrading performance, while a balanced dropout probability effectively suppresses noise in deeper timesteps while preserving critical information in shallower steps. This enables the network to navigate critical learning stages more effectively, enhancing both stability and robustness.



\paragraph{\textbf{Combinations of sub-modules in STAA with TSRD.}}
To evaluate the contribution of each sub-module in STAA, as well as the impact of TSRD, we conduct an ablation study by progressively integrating these components into the SNN. 
The results on CIFAR-100 are presented in \cref{tab:comb_staa}.
Notably, the GC block alone improves accuracy by 0.92\%, while the SA block contributes an additional 1.14\%, highlighting the importance of spatio-temporal attention and feature aggregation reconstruction in enhancing SNN performance. 
When all components, including TSRD, are incorporated, the accuracy reaches 75.10\%, demonstrating a cumulative improvement of 2.80\% over the baseline.

\paragraph{Impact of Intermediate Dimension Scaling Coefficients in GC.}
The GC module incorporates multiple \(1 \times 1\) convolutions for feature transformation.
Retaining the original channel size \(C\) for the last two convolutions layers would significantly increase computational overhead.
To maintain the module’s efficiency while preserving performance, a scaling coefficient \(r\) is introduced within the GC to compress feature dimensions.
By applying this scaling mechanism, the parameter count is reduced from \((C\cdot C)\) to \((2 C \cdot C / r)\), making the module more computationally feasible.
We conduct experiments to assess the impact of different \(r\) values, as shown in \cref{fig:scale}. The results indicate that the model achieves optimal performance when \(r = 4\), effectively balancing computational efficiency and accuracy.

\subsection{Visualization and Analysis}
\label{sec:exp_visual}

We aggregate 4D (\(T,C,H,W\)) spiking maps into 2D (\(H,W\)) and apply Grad-CAM to visualize feature extraction at different depths.
As shown in \cref{fig:heatmap}, the left displays a sample from the horse class in  CIFAR10-DVS, while the right compares vanilla LIF and STAA-LIF feature maps. STAA-LIF focuses more effectively on target-relevant features, capturing finer details like the horse’s mouth, which the vanilla LIF model misses. This highlights the advantage of spatio-temporal attention in enhancing feature selectivity.


\subsection{Analysis of Computation Efficiency}\label{sec:exp_energy}

We evaluate the single-image inference energy cost using 45-nm technology. ANNs rely on multiplication-and-accumulation (MAC) operations, which consume 4.6 pJ per operation, whereas SNNs primarily use accumulation (AC) operations at a significantly lower cost of 0.9 pJ per operation, making SNNs inherently more energy-efficient~\cite{horowitz_11_2014, qiao_reconfigurable_2015}.
Following \cite{chen_training_2023}, we assess the efficiency of ResNet-20, which requires 0.10 billion ACs, 0.06 billion MACs, 0.87 billion FLOPs, and consists of 12.69 million parameters. Under these conditions, inference on CIFAR-100 (with a 32×32 resolution and T=4) results in an energy consumption of 0.366 mJ, demonstrating the energy-efficient nature of ResNet-20. Additional computational efficiency experiments are provided in the Supplementary Material.

\section{Conclusion}
In this paper, we introduce STAA-SNN, a spatio-temporal attention-based feature aggregator, together with the TSRD training strategy. 
STAA-SNN leverages the inherent ability of SNNs to process spatio-temporal information, thereby enhancing network performance while reducing the number of required timesteps. Our approach achieves state-of-the-art results across multiple benchmarks.
STAA-SNN is designed as a plug-and-play module that seamlessly integrates into convolution-based SNN architectures. It is composed of several synergistic modules that collectively enhance its capabilities. The GC module simulates the brain’s ability to process information across a broad spatial range, involving multiple brain regions in coordination. By focusing on spatial feature relevance, it enables the model to capture global structures and key stimuli in its environment. The PE module enhances temporal encoding by incorporating time-step information into feature representations, mirroring the brain’s ability to integrate temporal and spatial cues for improved understanding of time-dependent relationships. 
The SA module mimics the brain’s dynamic attention mechanisms, adjusting feature importance across time steps by adaptively enhancing or suppressing information. This prioritization of critical temporal features strengthens decision-making and overall network efficiency.
Together, these modules form a robust spatio-temporal feature enhancement system that improves cognitive processing, robustness, and overall model performance. 
Despite these promising results, future research should focus on reducing computational energy consumption to further improve efficiency, particularly for resource-constrained applications.

\section{Acknowledgements}
This work was supported in part by National Natural Science Foundation of China (NSFC) (62476035, 62206037, 62276230, 62271361), and Natural Science Foundation of Zhejiang Province (LDT23F02023F02), and State Key Laboratory (SKL) of Biobased Transportation Fuel Technology.
{
    \small
    \bibliographystyle{ieeenat_fullname}
    \bibliography{main}
}
\clearpage
\setcounter{page}{1}
\maketitlesupplementary

\section{Datasets Details and Augmentation}

\paragraph{CIFAR-10.}
The CIFAR-10 dataset serves as a widely recognized benchmark for images, all with dimensions of $32\times32$.
This dataset contains 10 distinct classes,
encompassing various common objects such as airplanes, cars, birds, and cats.
It is frequently used as a standard for assessing the effectiveness of image classification algorithms, presenting a diverse array of visual challenges.
In our approach, we implement data augmentation techniques, including cropping, horizontal flipping, and cutout, throughout the entire training dataset.
In addition, during the training phase, we introduce random augmentation by selecting two strategies from the contrast enhancement, rotation, and translation.
These adjustments add robustness to the model and enhance its ability to generalize to diverse visual scenarios.

\paragraph{CIFAR-100}
CIFAR-100 \cite{krizhevsky2010cifar} represents an extension of the CIFAR-10 dataset and is designed to address more complex classification tasks.
It comprises 50,000 training images and 10,000 test images, all standardized at \(32 \times 32\) dimensions.
The dataset comprises 100 classes, each of which belongs to one of the 20 superclasses.
CIFAR-100 offers a more demanding challenge compared to CIFAR-10, making it a more suitable benchmark for evaluating the effectiveness of models in classification tasks.
The data augmentation strategy employed for CIFAR-100 aligns with that of CIFAR-10.

\paragraph{ImageNet.}
ImageNet contains a training dataset of 1.3 million images across 1,000 categories, along with an additional 50,000 images for validation.
Compared to the CIFAR-10/100 datasets, ImageNet presents a larger and more complex collection of images, providing a more robust benchmark for evaluating model generalization and learning capabilities.
In our experiments, we employ the data augmentation techniques outlined in \cite{he2016deep}.
Images are randomly cropped from either their original version or a horizontally flipped version to a size of $224 \times 224$ pixels, followed by data normalization.
For testing samples, images are resized to $224 \times 224$ pixels and subject to center cropping, after which data normalization is also applied.

\paragraph{CIFAR10-DVS.}
CIFAR10-DVS, introduced in \cite{li2017cifar10}, represents one of the largest visual neuromorphic datasets currently available.
It comprises 10,000 event streams, each with a size of \(128 \times 128\), derived from the frame-based CIFAR-10 images using a dynamic vision sensor (DVS).
The dataset comprises 10 categories, each containing 1,000 images.
During the training phase, the dataset is divided into training and testing datasets at a ratio of 9:1.
In the preprocessing stage, the training dataset undergoes random horizontal flipping, followed by the random selection of an augmentation technique such as cropping, translation, rotation, cutout, or erasing.
These techniques are used to enhance the diversity of the training dataset and strengthen the model's generalization capabilities\cite{10.1145/3503161.3547788, Guan_2020_CVPR}.



\paragraph{DVS128 Gesture.}
The DVS128 Gesture dataset, as presented in \cite{amir2017low}, is specifically curated for gesture recognition tasks.
It comprises 1,176 training images and 288 testing images, each with dimensions of \(128 \times 128\).
The dataset features 11 different gestures performed by 29 subjects under 3 illumination conditions, adding complexity to the recognition task.
This dataset serves as a valuable resource for evaluating models designed for gesture recognition in dynamic and varying conditions.
To enhance the dataset, each frame undergoes cutout and mixup operations, and random augmentations such as rotation, shear, and translation are utilized.
These techniques aim to enrich the dataset and improve the model's capacity to generalize across various gesture recognition scenarios.

\section{Experimental Settings}
All code implementations are based on the PyTorch framework.
Experiments were conducted on a single RTX 3090 GPU for all datasets except ImageNet, which was trained using a configuration of eight RTX 4090 GPUs.
In the experimental setup, we utilized the SGD optimizer with a momentum of 0.9 across all datasets and employed the CosineAnnealing learning rate adjustment strategy.

\paragraph{CIFAR-10/100.}
For CIFAR-10/100, we configured the initial learning rate to 0.1, batch size to 128, number of training epochs to 500, \(\alpha\) to 2, and \(\beta\) to 0.
Additionally, the dimension scaling factor of the hidden representations of \(W_q\) in the GC is set to 4, while the dimension scaling factor for the hidden representations in the SA is set to 16.

\paragraph{CIFAR10-DVS.}
In the case of CIFAR10-DVS, the initial learning rate is set to 0.01, batch size to 8, and the number of training epochs to 200.
Furthermore, \(\alpha\) is set to 2, and \(\beta\) to 0.1.
The scaling factor of the dimension size of the intermediate representations in the GC is set to 4, and in the SA it is set to 16.

\paragraph{ImageNet.}
For the ImageNet dataset, the initial learning rate is set to 0.1, with a batch size of 64 and a total of 350 training epochs.
Furthermore, the parameters \(\alpha\) and \(\beta\) are configured to 2 and 0.1, respectively.
The scaling factors for the intermediate dimension in GC and SA are set to 4 and 16, respectively.

\paragraph{DVS128 Gesture.}
For the DVS128 Gesture dataset, the learning rate is initialized at 0.01, and the batch size is set to 8.
The model undergoes 500 training epochs, with \(\alpha\) and \(\beta\) values set to 2 and 0.1, respectively.
The scaling coefficients for the two intermediate dimensions in GC and SA are set to 4 and 16.

\section{Supplemental Experiments}
\subsection{Impact of Intermediate Dimension Scaling Coefficients in GC.}
The GC module incorporates multiple \(1 \times 1\) convolutions.
Keeping the number of channels \(C\) consistent for the last two convolutions within the module would notably escalate computational costs.
To ensure the lightweightness of the module, a scaling coefficient \(r\) is incorporated in the GC to compress features.
This efficient approach reduces the module's parameter count from \(C \cdot C\) to \(2 \cdot C \cdot C / r\).

Evaluation involving different values of \(r\) is conducted on the ResNet-20 architecture  using a time step of 4 on the CIFAR-10 dataset.
The corresponding test accuracies for varied \(r\) values are detailed in Table \ref{tab:abla_r}, revealing optimal performance when \(r\) is set to 4.

\begin{table}[h]
\centering
\begin{tabular}{cccc}
\toprule
Dataset                   & Architecture               & r  & Accuracy \\ 
\midrule
\multirow{5}{*}{CIFAR-10} & \multirow{5}{*}{ResNet-20} & 1  & 94.93\%  \\
                          &                            & 2  & 94.85\%  \\
                          &                            & 4  & 95.03\%  \\
                          &                            & 8  & 94.81\%  \\
                          &                            & 16 & 94.72\%  \\ 
\bottomrule
\end{tabular}
\caption{Impact of different scaling coefficients \(r\) for intermediate feature dimensions in the GC module on the CIFAR-10 dataset.}
\label{tab:abla_r}
\end{table}

\begin{table}[h]
\centering
\begin{tabular}{ccc}
\toprule
Pooling Method                          & Timestep & Accuracy \\ \midrule
\multirow{2}{*}{+1 MaxPool}             & 1        & -        \\
                                        & 2        & 93.86\%  \\ \midrule
\multirow{2}{*}{+1 AvgPool}             & 1        & 91.27\%  \\
                                        & 2        & 93.96\%  \\ \midrule
\multirow{2}{*}{+1 AvgPool, +1 MaxPool} & 1        & 89.60\%  \\
                                        & 2        & 94.44\%  \\ \midrule
\multirow{2}{*}{+2 AvgPool}             & 1        & 92.91\%  \\
                                        & 2        & 94.27\%  \\ \bottomrule
\end{tabular}
\caption{Ablation Study of Pooling Method Combinations in SA with ResNet-20 on CIFAR-10.}
\label{tab:alb_pool}
\end{table}

\subsection{Pooling Method Selection in SA.}
Evaluation of various pooling methods for image classification was conducted, and the outcomes are detailed in Table \ref{tab:alb_pool}.
The analysis revealed that the overall performance was relatively better, and the network could converge effectively when utilizing average pooling.
In contrast, the incorporation of max pooling posed significant challenges, including difficulties in training and the potential failure to converge.
Moreover, the utilization of max pooling substantially extended the training time.

Further experimentation involving different combinations of average pooling demonstrated that the use of two average pooling layers yielded superior performance compared to a single layer.
This observation validates the integration of the \(\alpha\) parameter in SA, set to 2.
Consequently, to balance both accuracy and training speed, our final selection involved a combination scheme utilizing multiple average pooling layers.

\begin{table*}[t]
\centering
\begin{tabular}{|c|c|c|c|c|c|c|c|}
\hline
\textbf{Arch.} & \textbf{Res.} & \textbf{T} & \textbf{ACs(G)} & \textbf{MACs(G)} & \textbf{FLOPs(G)} & \textbf{Params(M)} & \textbf{Energy(mJ)} \\ \hline
ResNet20              & 224x224             & 4          & 5.38            & 2.82             & 42.62             & 13.20             & 17.814             \\ \hline
VGG13                 & 224x224             & 4          & 3.45            & 13.56            & 45.07             & 11.17             & 65.481             \\ \hline
ResNet20              & 32x32               & 4          & 0.10            & 0.06             & 0.87              & 12.69             & 0.366              \\ \hline
VGG13                 & 32x32               & 4          & 0.05            & 0.28             & 0.92              & 10.67             & 1.333              \\ \hline
\end{tabular}
\caption{Table of computational consumption for different models on CIFAR-100 and ImageNet.}
\label{tab:cost}
\end{table*}

\section{Analysis of Computation Efficiency}
In ANNs, each operation involves a multiplication and accumulation (MAC) process. The total number of MAC operations (\#MAC) in an ANN can be calculated directly and remains constant for a given network structure. In contrast, spiking neural networks (SNNs) perform only an accumulation computation (AC) per operation, which occurs when an incoming spike is received. The number of AC operations can be estimated by taking the layer-wise product and sum of the average spike activities, in relation to the number of synaptic connections.

\begin{equation}
\left\{
\begin{aligned}
\# &\text{MAC} = \sum_{l=1}^{L} (\# \text{MAC}_{l}) \\
\# &\text{AC} = \sum_{l=2}^{L} (\# \text{MAC}_{l} \times a_{l}) \times T
\end{aligned}
\right.
\end{equation}
Here, $a_{l}$ represents the average spiking activity for layer l. The first, rate-encoding layer of an SNN does not benefit from multiplication-free operations and therefore involves MACs, while the subsequent layers rely on ACs for computation.

The energy consumption E for both ANN and SNN, accounting for MACs and ACs across all network layers, is given by:

\begin{equation}
\left\{
\begin{aligned}
E_{SNN} &=  \# \text{MAC}_{1}  \times E_{MAC}+ \# \text{AC} \times E_{AC} \\
E_{ANN} &= \# \text{MAC} \cdot E_{MAC}
\end{aligned}
\right.
\end{equation}

Based on previous studies in SNN research \cite{yao_attention_2023,chakraborty_fully_2021}, we assume that the operations are implemented using 32-bit floating-point (FL) on a 45 nm 0.9V chip \cite{horowitz_11_2014}, where a MAC operation consumes 4.6 pJ and an AC operation consumes 0.9 pJ. This comparison suggests that one synaptic operation in an ANN is roughly equivalent to five synaptic operations in an SNN. It is important to note that this estimation is conservative, and the energy consumption of SNNs on specialized hardware designs can be significantly lower, potentially reduced by up to 12× to 77 fJ per synaptic operation (SOP) \cite{qiao_reconfigurable_2015}.
We conduct energy consumption tests on different models using CIFAR-100 and ImageNet dataset, and the specific results are recorded in Table \ref{tab:cost}.


\end{document}